\documentclass[10pt,twocolumn,letterpaper]{article}

\usepackage{isba}
\usepackage{times}
\usepackage{epsfig}
\usepackage{graphicx}
\usepackage{amsmath}
\usepackage{amssymb}

\usepackage{multirow}
\usepackage{textcomp}
\usepackage{array}

\usepackage{url}

\isbafinalcopy 

\ifisbafinal\fi

\usepackage{fancyhdr}
\begin{document}

\title{Iris Recognition with a Database of Iris Images Obtained in \\Visible Light Using Smartphone Camera}

\author{Mateusz Trokielewicz$^{1,2}$\\
$^{1}$Biometrics Laboratory\\
Research and Academic Computer Network\\
Wawozowa 18, 02-796 Warsaw, Poland\\
$^{2}$Institute of Control and Computation Engineering\\
Warsaw University of Technology\\
Nowowiejska 15/19, 00-665 Warsaw, Poland\\
{\tt\small mateusz.trokielewicz@nask.pl}}

\maketitle
\thispagestyle{empty}

\begin{abstract}
This paper delivers a new database of iris images collected in visible light using a mobile phone's camera and presents results of experiments involving existing commercial and open-source iris recognition methods, namely: IriCore, VeriEye, MIRLIN and OSIRIS. Several important observations are made. 

First, we manage to show that after simple preprocessing, such images offer good visibility of iris texture even in heavily-pigmented irides. Second, for all four methods, the enrollment stage is not much affected by the fact that different type of data is used as input. This translates to zero or close-to-zero Failure To Enroll, i.e., cases when templates could not be extracted from the samples. Third, we achieved good matching accuracy, with correct genuine match rate exceeding 94.5\% for all four methods, while simultaneously being able to maintain zero false match rate in every case. Correct genuine match rate of over 99.5\% was achieved using one of the commercial methods, showing that such images can be used with the existing biometric solutions with minimum additional effort required. Finally, the experiments revealed that incorrect image segmentation is the most prevalent cause of recognition accuracy decrease.

To our best knowledge, this is the first database of iris images captured using a mobile device, in which image quality exceeds this of a near-infrared illuminated iris images, as defined in ISO/IEC 19794-6 and 29794-6 documents. This database will be publicly available to all researchers.

\end{abstract}

\section{Introduction}
Most of the existing non-mobile iris biometric approaches take advantage of images collected in near-infrared (NIR) illumination, that is supposed to offer better visibility of iris texture, especially for heavily pigmented irides. Such setup, however, is much more difficult to implement in portable devices, such as phones and tablets. As of today, with the notable exception of Fujitsu NX F-04G near infrared iris recognition-equipped smartphone (\cite{fujitsuIrisPhone}, sold exclusively in Japan), this usually requires an additional device that connects to a phone or a tablet. With additional cost, low practicability and limited availability of SDKs for mobile operating systems, it is not an easy task to popularize such solutions. On the other hand, if good accuracy could be achieved with images obtained using cameras already embedded in consumer devices, an opportunity for low-effort implementations would arise. 

When investigating past work on the matter (Sec. \ref{section:related}), we find that most of the studies are devoted to dealing with very poor quality images obtained using mobile devices, usually in very unconstrained conditions (low light, shadows, off-angle gaze direction, motion blur, out of focus images). Despite the fact that there are several databases of such images available to the public, none of them offers an extensive collection of good quality images.

However, mobile devices has recently become more and more advanced, and the progress applies to photo-taking capabilities as well. Therefore, this study presents a different approach, focusing on acquiring good quality samples, rather than deploying advanced methodologies for coping with negative effects introduced by the poor quality ones. \textbf{The following four questions are considered in this paper:}

\textbf{Question 1:} Do iris images captured with a mobile phone, after non-computationally-extensive preprocessing, offer sufficient visibility of iris texture details for all levels of pigmentation?

\textbf{Question 2:} Can visible light iris images be successfully enrolled with existing, iris recognition algorithms that were originally developed with NIR images in mind?

\textbf{Question 3:} Is the verification accuracy using such algorithms sufficient to establish a working biometric system?

\textbf{Question 4:} If recognition errors are observed, what are the reasons for non-ideal behavior? 

To provide answers to those questions, an appropriate database of iris images taken with iPhone 5s' rear camera has been collected (Sec. \ref{section:database}). Experimental study regarding performance assessment done for four different, commercial and academic algorithms is presented in Sec. \ref{section:experiments}. To our best knowledge, this is the only dataset of high quality iris images obtained using a mobile phone's camera with embedded flash, and the only study that aims at using this kind of images with existing iris recognition algorithms. Notably, this study also contributes to the field of iris biometrics by making the dataset publicly available to the research community for non-commercial purposes. 

\begin{figure*}[!t]
\centering
\includegraphics[width=0.249\textwidth]{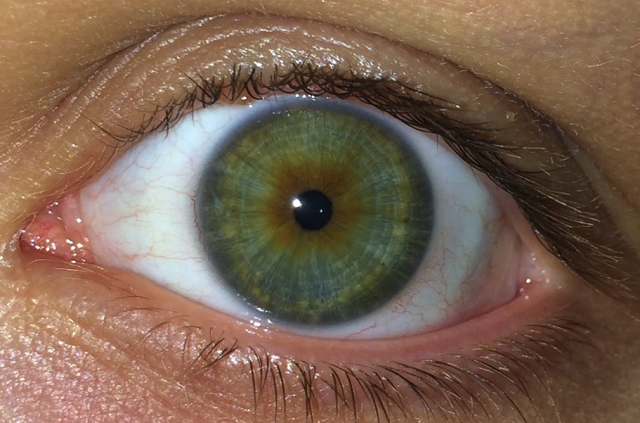}\hfill
\includegraphics[width=0.249\textwidth]{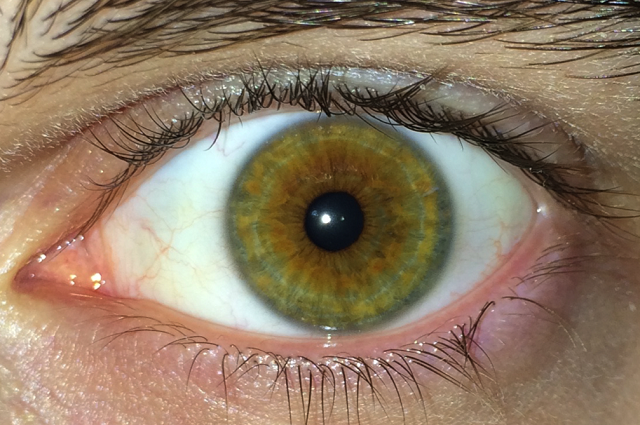}\hfill
\includegraphics[width=0.249\textwidth]{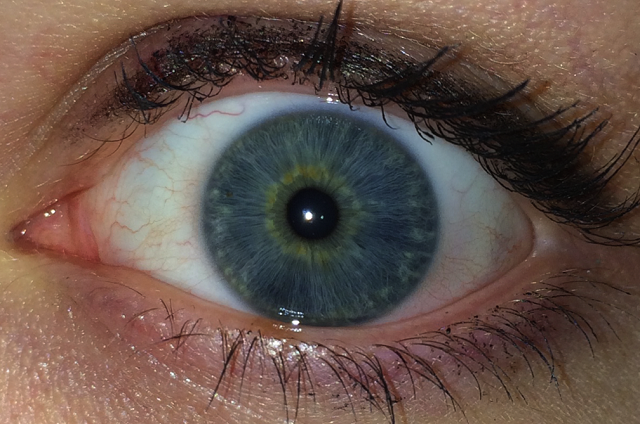}\hfill
\includegraphics[width=0.249\textwidth]{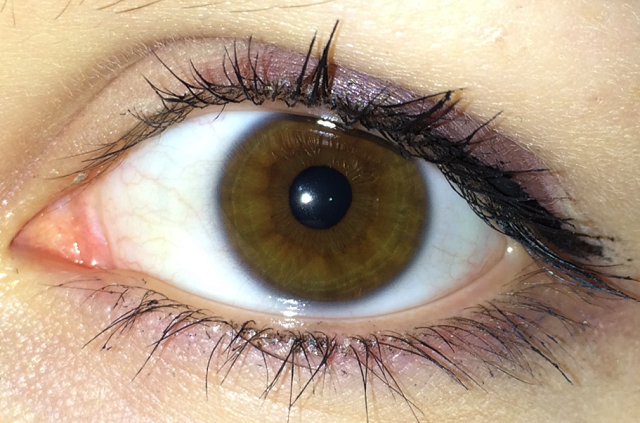}
\vfill
\includegraphics[width=0.249\textwidth]{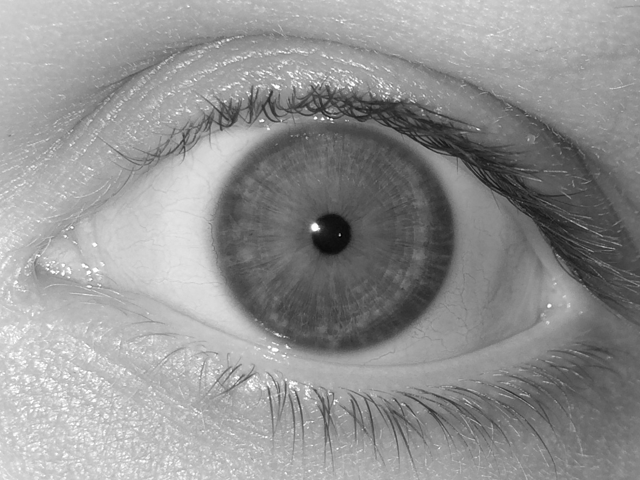}\hfill
\includegraphics[width=0.249\textwidth]{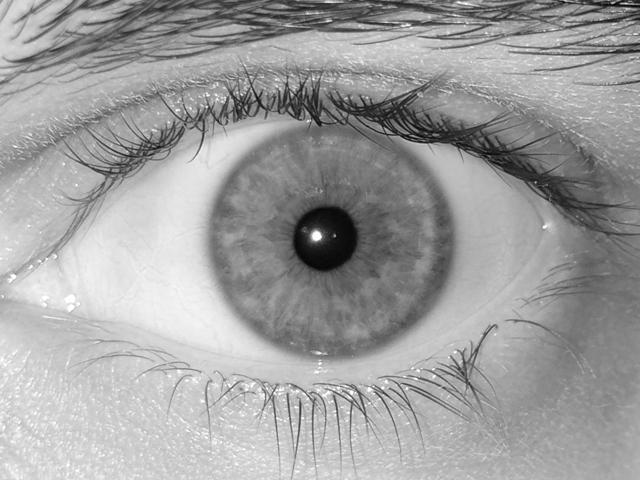}\hfill
\includegraphics[width=0.249\textwidth]{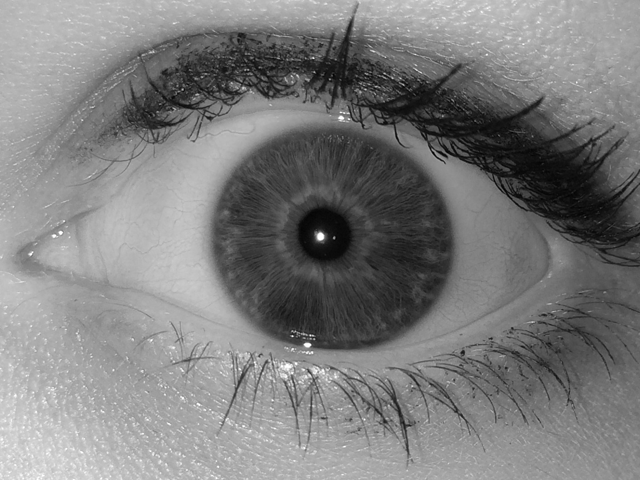}\hfill
\includegraphics[width=0.249\textwidth]{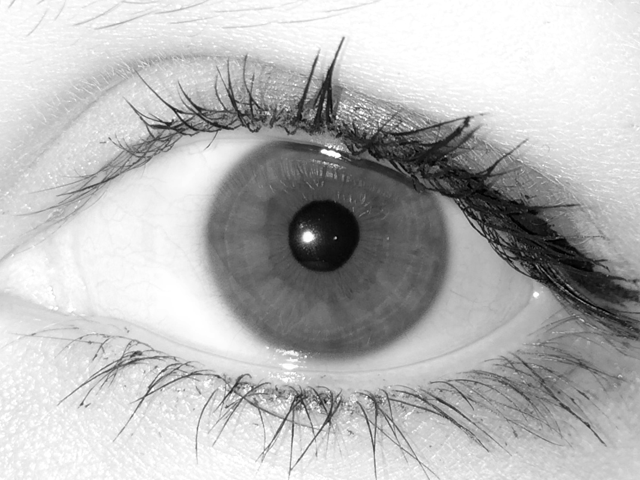}
\vskip0.2cm
\caption{Cropped visible spectrum photographs obtained using the iPhone 5s and corresponding images after grayscale conversion for four different types of pigmentation (\textbf{from left to right}): light blue/grey, hazel/green, dark blue, dark brown. Using the red channel of the RGB color space enables good visibility of the iris pattern even in the heavily pigmented irides.}
\label{fig:samples_converted}
\end{figure*}

\section{Related work}
\label{section:related}
First publicly available databases to present iris images obtained in visible light were the UPOL \cite{UPOLCzechy2004}, UBIRISv1 \cite{UBIRISv1} and UBIRISv2 \cite{UBIRISv2} datasets. The first one contains images collected with an ophthalmology device, the second - scaled down photographs taken with Nikon E5700 handheld camera (resampling was introduced to simulated poor quality of unconstrained imaging), and the third - images captured \emph{on-the-move} and \emph{at-a-distance}, also in visible light. Recently, a dataset of iris images and iris image printouts (`fake' iris images) obtained using various mobile devices has been released \cite{MICHE2015}, however, these are also images taken without built-in flash illumination.  

Upon the UBIRISv2 database, the NICE.I competition was founded by Proenca \etal \cite{NICE1}, focusing on the noisy iris images and independent evaluation of segmentation and noise recognition stages, as those two are acknowledged by the authors to be the most likely source of errors. Following that, Proenca discusses challenges associated with unconstrained iris recognition in visible light \cite{ProencaFeasibility}, most notably the amount of information that can be captured using visible light acquisition setups. Santos \etal explore configurations of visible light illumination best for unconstrained iris recognition \cite{SantosPreliminaryAssessment}, such as the choice of illuminant and its luminance in respect to different eye pigmentation types. Proenca also proposes a new method for segmenting poor quality iris images captured in visible illumination \cite{ProencaVisibleSegmentation}, and methods for assessing image quality in order to improve an overall system performance by discarding samples of substandard quality \cite{ProencaVisibleQuality}.

Raja \emph{et al.} investigate visible light iris recognition using a light field camera \cite{KiranLightField2013}, and also smartphones and tablets \cite{KiranSparse2014}, reporting promising results. In the latter scenario, EER (Equal Error Rate) below 2\%  is achieved, for a method involving improved OSIRIS segmentation and feature extraction algorithm incorporating deep sparse filtering. They also explore K-means clustering as a possible recognition technique for visible light images with best achieved EER of only 0.31\% \cite{KiranKmeans}. Recently, these researchers also investigated a new setup for capturing iris images in white LED illumination using mobile devices \cite{KiranBTAS2015}, achieving recognition accuracy of up to 90\% GMR (Genuine Match Rate) with 0.1\% FMR (False Match Rate) using a combination of Daugman method and images captured with Nokia Lumia 1020 device. Our own previous experiments devoted to the field of visible light iris recognition were presented in \cite{TrokielewiczWilga2015}. We managed to achieve EERs not exceeding 8\% for two commercial iris recognition methods with iris images collected using a smartphone camera. 

Segmentation stage is recognized as the most challenging part of the recognition process using noisy images in a study by Radu \emph{et al.} \cite{RaduColorIrisRecognition2013}, where authors manage to achieve 7\% segmentation error rate and 3.7\% classification EER on the UBIRISv1 dataset. Difficulties with segmenting poor quality iris images are also reported by Frucci \emph{et al.} \cite{IDEM2014}, who developed a segmentation method aimed specifically at noisy iris images that is said to outperform other approaches (MICHE database was used for calculations). Possibility of deploying iris and face biometrics onto mobile devices is also investigated by De Marsico \emph{et al.} \cite{FIRMEfaceIrisApp2014}.

\section{Database of visible light iris images}
\label{section:database}
A new dataset of iris images obtained in visible light using a mobile phone has been designed and collected for the purpose of this study. All volunteers participating in the experiments have been provided with information regarding this research and an informed consent has been obtained from each participant. 

70 people participated in the data collection. For each volunteer, the data were collected during two acquisition sessions that were 3 to 7 days apart, depending on a given person's availability. Apple iPhone 5s has been chosen as the capturing device due to its high quality 8 megapixel sensor and a large lens aperture of \emph{f/2.2}. Only the rear camera has been used and flash was enabled during all sample acquisitions. This is to ensure that all images in the database represent a visible, well-lit iris regardless of the natural lighting conditions. Data collection was performed indoors. After each sample acquisition the subject was asked to look away from the camera and blink a few times to deliberately introduce some level of intra-session noise, so that the samples are more likely to resemble those in a real-world use case (such as authenticating the user for phone access). 
\pagebreak

The final dataset comprises \textbf{3192 color images of 139 and 136 different irides}, acquired in sessions 1 and 2, respectively. See Fig. \ref{fig:samples_converted} (top row) for sample images from the database. 

\section{Experimental study}
\label{section:experiments}

\subsection{Iris recognition algorithms}
Four different, commercial and academic solutions for iris recognitions have been used for the purpose of this experiment. \textbf{VeriEye} \cite{VeriEye} employs an unpublished, proprietary algorithm that is only said to use non-circular approximation for pupillary and limbic boundaries of the iris and encoding different from Gabor-based image filtering. Comparison scores returned by this method utilize a non-standard score scale, starting from zero (when different eyes are compared) to some, unknown values for same-eye comparisons. \textbf{MIRLIN}\cite{MIRLIN}\cite{Monro2007} derives iris features from zero-crossings of the differences between Discrete Cosine Transform calculated for overlapping, angular patches in the iris image. Comparison scores are calculated in the form of Hamming distance between the binary feature sets, resulting in scores that should be close to zero for same-eye comparisons and around 0.5 for different-eye comparisons. \textbf{OSIRIS} \cite{OSIRIS} is an open-source solution following a concept proposed by Daugman, that involved image normalization into a polar representation and Gabor-based filtering. Binary iris code is then compared, resulting in a match score in a form of Hamming distance, as in the MIRLIN method. Here as well, values close to zero are expected for same-eye, and close to 0.5 for different-eye comparisons. Similarly to the VeriEye method, \textbf{IriCore} \cite{IriCore} incorporates an unpublished method that returns the comparison score in a form of dissimilarity metric that is expected to be near zero for same-eye comparisons, while scores for different-eye images should be between 1.1 and 2.0.

\subsection{Evaluation methodology}
To answer \textbf{Question 1}, simple image preprocessing is performed on the raw data from the smartphone to achieve good visibility of the iris texture regardless of its color. Then, four iris recognition methods are used to enroll all samples, \emph{i.e.,} create iris templates - mathematical representations of information found in the iris patterns. FTEs (Failure-To-Enroll values) are calculated to answer \textbf{Question 2}. Finally, full sets of genuine (same-eye) comparisons, as well as impostor (different-eye) comparisons are performed to calculate similarity or dissimilarity scores and come up with ROC curves and CMRs (Correct Match Rate values) and answer \textbf{Question 3}. Every possible pair of images is used only once for matching, \emph{i.e.,} if image A is compared against image B, then image B is not compared against image A (as all four methods employed in this study return the exact same comparison score in either scenario). Finally, we employ one of the iris recognition methods to return image segmentation results and check whether erroneous iris localization contributes to the non-ideal recognition accuracy (\textbf{Question 4}).

\begin{figure*}[!t]
\centering
\includegraphics[width=\textwidth]{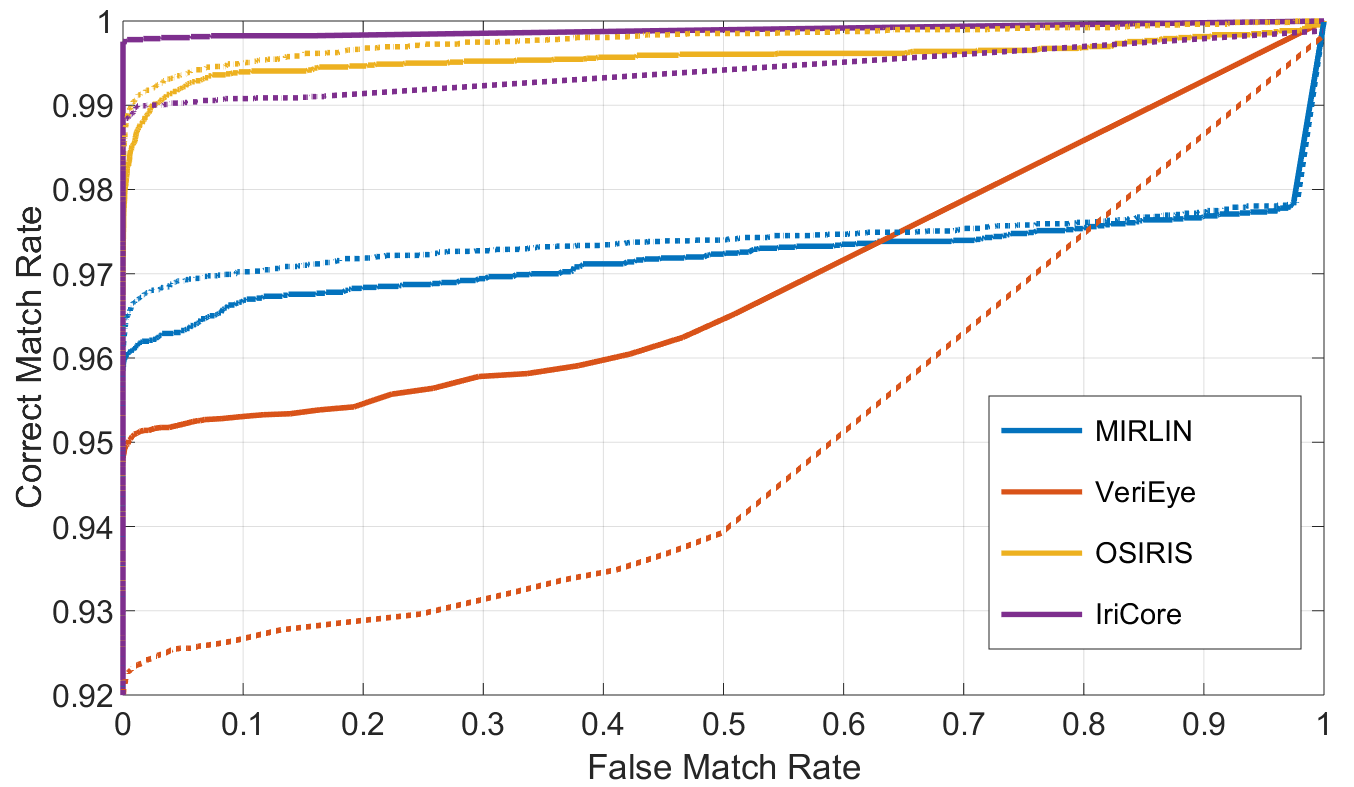}
\vskip0.2cm
\caption{Receiver operating characteristics (ROCs) for MIRLIN, VeriEye, OSIRIS and IriCore methods, calculated from scores obtained using images from the first (\textbf{solid lines}) and the second (\textbf{dotted lines}) acquisition session.}
\label{fig:ROC_session1}
\end{figure*} 

\subsection{Image preprocessing}
Typical iris recognition cameras usually perform certain image preprocessing after taking the photograph to make the resulting file compliant with the ISO/IEC regulations defining constraints for iris images \cite{ISO}\cite{ISO2}. The most important requirements define the iris image as an 8-bit grayscale VGA bitmap with predefined margins and proportions. This is to ensure preliminary compatibility with iris recognition algorithms. However, when working with images taken with a mobile phone, it is researcher's responsibility to secure these requirements, if such images are intended to be used with existing algorithms.

Therefore, a certain image preprocessing had to be performed. This was done manually, as our intention was to leverage effects originating in sample image differences (visible light vs NIR), and not to build a robust segmentation algorithm. Operations include cropping original images to the VGA resolution with iris centered and grayscale conversion using the red channel. As wavelengths corresponding to red light are the longest in visible spectrum (and thus closest to near infrared), we expect to get the best iris pattern visibility this way. This approach lets us build a dataset populated with samples that generally resemble those coming from a typical NIR system, but represent different data. Fig. \ref{fig:samples_converted} shows cropped color images and corresponding grayscale images. Easily discernible details of the iris pattern regardless of pigmentation levels allow us to state that \textbf{the answer to Questions 1 is affirmative, as even raw images offer quality that highly exceeds ISO/IEC requirements, and after subsequent preprocessing good visibility of the iris tissue even for highly-pigmented irides is easily achieved.} 

\subsection{Enrollment performance}

FTE rates achieved when attempting to enroll images into each one of four algorithms are shown in Table \ref{table:FTE}. Enrollment performance of all four methods is very satisfactory, with FTE not exceeding 1.27\%, and equaling zero or close-to-zero in most combinations of algorithm and acquisition session. \textbf{Thus, the answer to Question 2 is also affirmative, as no apparent algorithm misbehavior can be observed at this point.}

\begin{table}[!ht]
\renewcommand{\arraystretch}{1.1}
\caption{FTE rates obtained for each session and each algorithm.}
\vskip0.2cm
\label{table:FTE}
\centering\footnotesize
\begin{tabular}[t]{|c|c|c|c|c|}
\hline
\textbf{Acquisition session} & {VeriEye} & {MIRLIN} & OSIRIS & IriCore \\
\hline
\hline
\textbf{First} & 0\% & 1.27\%  & 0\% & 0\% \\
\hline
\textbf{Second} & 0.11\% & 1.12\%  & 0.11\% & 0\% \\
\hline
\end{tabular}
\end{table}

\subsection{Matching performance}

The accuracy of matching stage is shown in Fig. \ref{fig:ROC_session1} with receiver operating characteristic (ROC) curves plotted collectively for every employed algorithm to easily compare the performance that can be achieved with this database.

Correct Match Rate values have been calculated at the acceptance threshold that allows to reach $FMR=0\%$ (referred to as \emph{CMR@zeroFMR}) while using our dataset, see Table \ref{table:CMR}. All four methods behave well with the visible light data, maintaining good performance with CMR@zeroFMR not dropping below 91.93\% and reaching more than 99.5\% and 98.5\% with session 1 and session 2 data, respectively.

Based on the good performance of all methods we used in this experiment, we state that \textbf{the answer to Question 3 is affirmative, as iris biometric systems can be deployed using samples acquired in visible light}.

\begin{table}[!ht]
\renewcommand{\arraystretch}{1.1}
\caption{CMR@zeroFMR values obtained for each algorithm and acquisition session.}
\vskip0.2cm
\label{table:CMR}
\centering\footnotesize
\begin{tabular}[t]{|c|c|c|c|c|}
\hline
\textbf{Acquisition session} & {VeriEye} & {MIRLIN} & OSIRIS & IriCore \\
\hline
\hline
\textbf{First} & 94.57\% & 95.63\%  & 95.25\% & \textbf{99.67\%} \\
\hline
\textbf{Second} & 91.93\% & 96.02\%  & 97.87\% & \textbf{98.82\%} \\
\hline
\end{tabular}
\end{table}

\subsection{Sources of errors}
When looking at the shapes of the ROCs, it can be noticed that for the VeriEye matcher and especially the MIRLIN matcher there is a steep linear fragment of the curve. We assume that despite low FTE rates, some of the images were incorrectly segmented and the subsequent stage of iris image encoding is performed on the wrong (non-iris) data. As a result, some of the encoded data represents information that rather relates to eyelids, eyelashes or sclera portions than person-specific iris information used for authentication. 

To know if this is the case here, we have employed the MIRLIN matcher to generate images with iris localization results printed on the image itself. We then selected image pairs that yielded the worst same-eye comparison scores, as incorrect localization is usually the cause of increased (False Non-Match Rate), and not the increased FMR (False Match Rate). The criteria for choosing these pairs was to select those same-eye scores that were worse (\emph{i.e.,} higher, as Hamming distance is used as a dissimilarity metric) than the worst (\emph{i.e.,} lowest) different-eye scores. This returned a set of 190 image pairs that were subjected to manual visual inspection. This was done on the session 1 data only due to the time-consuming nature of the task.

\begin{figure}[!t]
\centering
\includegraphics[width=0.238\textwidth]{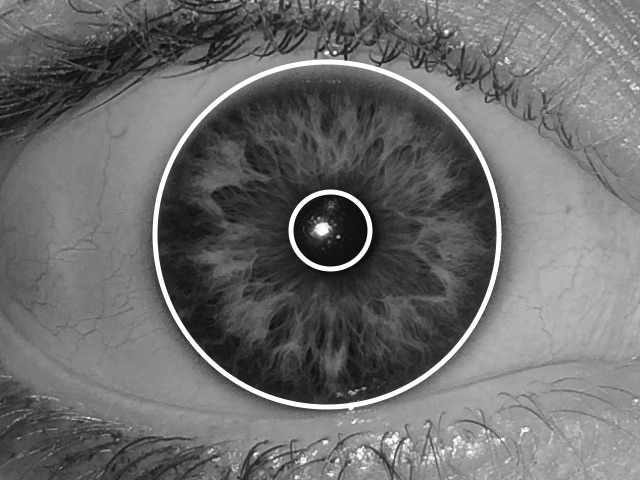}\hfill
\includegraphics[width=0.238\textwidth]{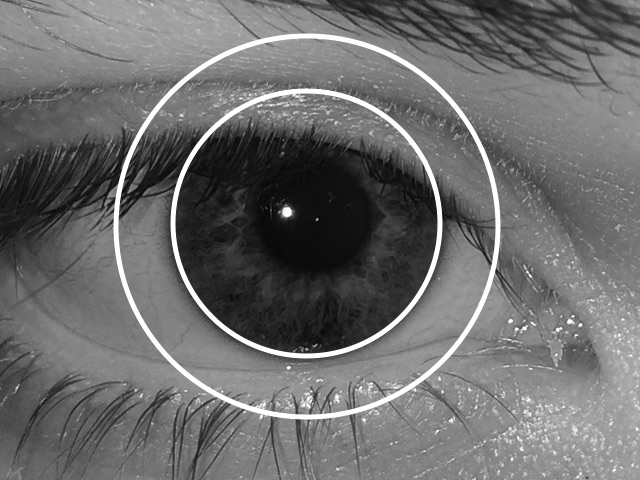}\vfill
\includegraphics[width=0.238\textwidth]{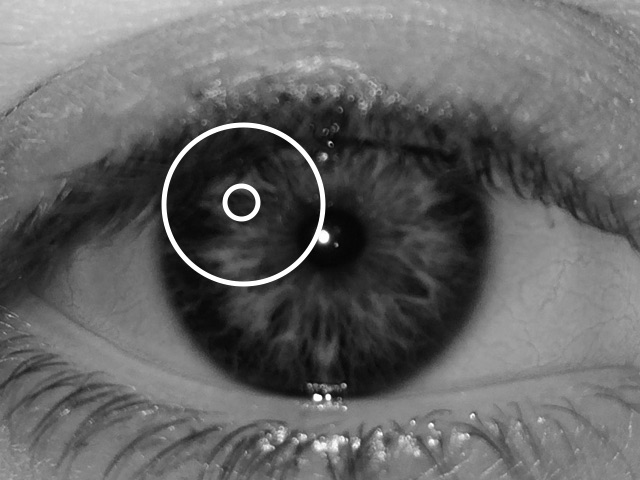}\hfill
\includegraphics[width=0.238\textwidth]{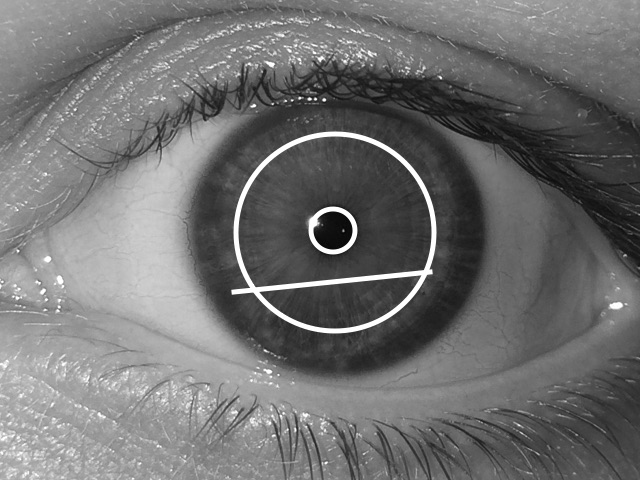}
\vskip0.2cm
\caption{Sample images with denoted results of the iris localization performed by MIRLIN algorithm. Correctly localized iris is represented in the first image in the upper left. The other three images present irides that were incorrectly localized.}
\label{fig:segmentation}
\end{figure}

Out of these 190 scores, \textbf{72.1\%} turned out to be affected by issues with image segmentation (selected cases are shown in Fig. \ref{fig:segmentation}). Other probable reasons for poor performance included excessive eyelid/eyelash occlusion (\textbf{15.3\%} of the scores) and blurred/out of focus images (\textbf{7.9\%}). The remaining \textbf{4.7\%} of the scores did not reveal any apparent flaw in either of the two images.

\textbf{Thus, the answer to the last, fourth question is the following: the majority of exceptionally bad comparison scores come from images with incorrect segmentation. Other reasons include blurred images (both due to motion blur and being out of focus) and excessive eyelid/eyelash occlusion.}

\section{Conclusions and future work}
This paper introduces a completely new database of high quality iris images obtained with a smartphone camera and provides answers to several questions regarding iris recognition accuracy with such data. Experiments involving four different, commercial and open-source iris recognition algorithms are performed to assess whether good quality visible light images are viable for use with existing iris recognition software without advanced preprocessing. By simply isolating the red channel in the RGB color space we have generated data that  yields good enrollment performance, with maximum Failure-To-Enroll (FTE) of only 1.27\%. In most cases, however, the sample FTEs were close or equal to zero. The subsequent stage - matching - also seems to be barely affected by differences in input data (using visible light spectrum instead of typical NIR), as we managed to achieve CMRs of more than 99.5\% for the zero FMR threshold.

Analysis devoted to finding the most prevalent sources of errors revealed that incorrect image segmentation is the most likely to cause a drop in recognition accuracy. This clearly shows that robust iris localization and image segmentation are crucial for achieving great iris recognition accuracy, as there are no obstacles for employing images captured in visible light in algorithms that were developed \emph{specifically} images captured in near infrared. 

With such promising results, future work is certainly necessary to fully explore the inherent potential of visible spectrum iris recognition. Our next steps will involve implementing one of the open source iris algorithms directly on the mobile device, together with fully automated iris localization and image preprocessing to bring the experiments much closer to a real-world scenario.

\section*{Acknowledgements}
The author would like to thank his colleagues - Ms Ewelina Bartuzi, Ms Kasia Michowska, Ms Tosia Andrzejewska and Ms Monika Selegrat from the Biometrics Scientific Club at the Warsaw University of Technology for their help with database collection and manual processing of database images. The author is also indebted to Dr Adam Czajka for his valuable insight and advice that contributed to the shape of this study. Finally, we would like to express our thanks to all of the volunteers that participated in the data collection process. 

{\small
\bibliographystyle{ieee}
\bibliography{refs}
}

\end{document}